\documentclass[times,twocolumn,final,authoryear]{elsarticle}

\usepackage{ycviu}
\usepackage{framed,multirow}

\usepackage{amssymb}
\usepackage{latexsym}

\usepackage{url}
\usepackage{xcolor}
\usepackage{amsmath}
\usepackage{booktabs} 
\usepackage{tabularx}
\usepackage{hyperref}
\definecolor{newcolor}{rgb}{.8,.349,.1}
\usepackage{appendix}

\journal{Computer Vision and Image Understanding}

\begin{document}

\clearpage
\thispagestyle{empty}
\ifpreprint
  \vspace*{-1pc}
\fi

\begin{frontmatter}

\title{VC-FeS: Viewpoint-Conditioned Feature Selection for Vehicle Re-identification in Thermal Vision}

\author[1]{Yasod \snm{Ginige}} 
\author[2]{Ransika \snm{Gunasekara}}
\author[2]{Darsha \snm{Hewavitharana}}
\author[2]{Manjula \snm{Ariyarathne}}
\author[2]{Peshala \snm{Jayasekara}}
\author[2]{Ranga \snm{Rodrigo}\corref{cor1}}
\cortext[cor1]{Corresponding author: 
   }
\ead{ranga@uom.lk}

\address[1]{School of Computer Science, University of Sydney, NSW (2008), Australia}
\address[2]{Department of Electronics and Telecommunication Engineering, University of Moratuwa, Katubedda, Colombo (10400), Sri Lanka}


\begin{abstract}
Identification of less-articulated objects using single-channel images, such as thermal images, is important in many applications, such as surveillance. However, in this domain, existing methods show poor performance due to high similarity among objects of the same category in the absence of color information (overlooking shape information) and de-emphasized texture information. Furthermore, variability in viewpoint adds more complexity as the features vary from side to side. We address these issues by constructing viewpoint-conditioned feature vectors and area-specific feature comparisons in separate feature spaces. These interventions enable leveraging the advancements of existing RGB-pre-trained ViT feature extractors while effectively adapting them to address the challenges specific to the thermal domain.
We test our system with RGBNT100 (IR) vehicle dataset and a thermal maritime dataset acquired by us. Our results surpass the state-of-the-art methods by 19.7\% and 12.8\% for the above datasets in mAP scores, respectively. 
We also plan to make our themal dataset available, the first of its kind for maritime vessel identification.



\end{abstract}


\end{frontmatter}


\section{Introduction}\label{sec1}
Re-identification (ReID)---the problem of matching objects across different camera views over time, despite variations in appearance due to changes in viewpoint---is the base problem for many use cases such as face recognition, human authentication, vehicle identification, and other surveillance systems ~(\cite{zahra2023person, shi2022iranet,shen2023git}). It is well explored in the RGB domain, and the existing methods provide promising results. One common solution for ReID in computer vision is to use networks such as ViT~(\cite{DOSOVI21}) or ResNet~(\cite{HEKAIM16}), to extract feature vectors from images and compare them using a distance metric to identify similar objects~(\cite{luo2019bag}). This has evolved in the recent past to re-identify objects from different viewpoints by paying more attention to color, texture patterns, and other common features that are visible from every viewpoint~(\cite{dong2024multi, SPAN}). 
Unfortunately, their performance severely degrades in the absence of color and texture, which is the very situation in thermal and low-light vision~(\cite{huang2023deep}).

\begin{figure*}[t]
    \centering
    \includegraphics[width=\linewidth]{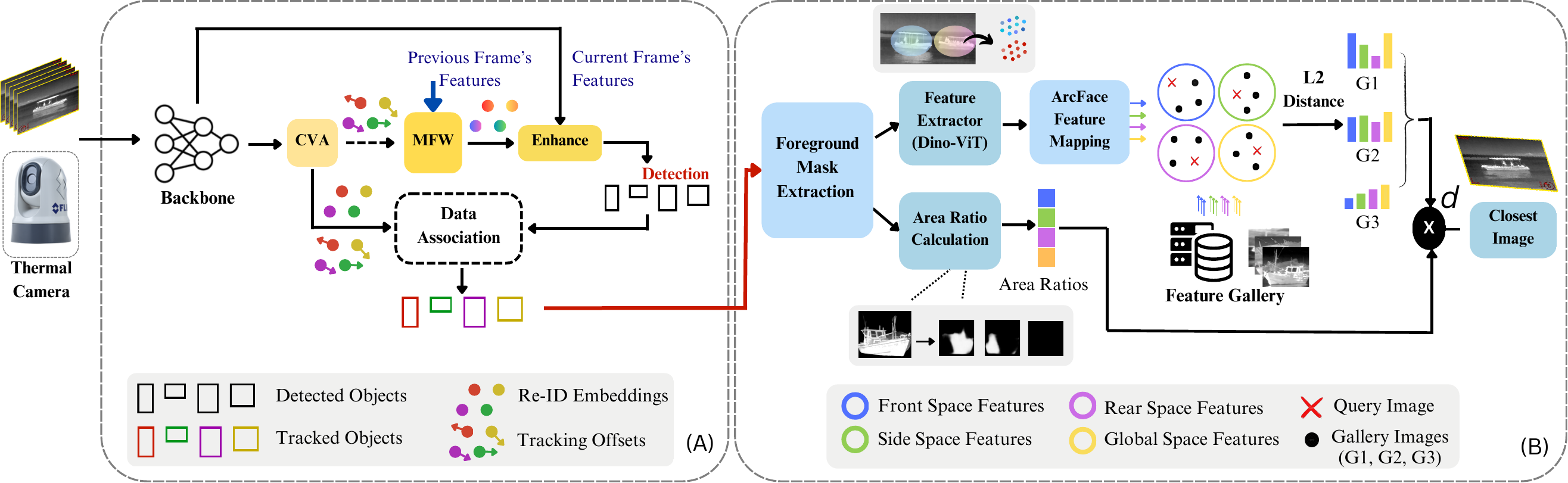}
    \caption{Overall structure of the proposed re-identification algorithm. Object detection and tracking algorithm (A) identifies objects by enhanced detection using the previous frame's features and tracking offsets. The detected objects are matched to relevant tracklets using offsets and ReID embeddings before being cropped and fed to the ReID algorithm (B), which extracts features from each visible side of the vehicle and compares them with the gallery. The feature comparison is done in four separate feature spaces and weighted using area ratios of each viewpoint to calculate the final distance matrix. Sections~\ref{Object_track} and~\ref{subsec3.2} further discuss (A) and (B) subsystems, respectively.}
    \label{big_picture}
\end{figure*}

The performance of ReID systems heavily depends on the features they extract.
In face recognition, existing methods lock on to rich facial landmarks such as distances between eyes, special lineaments, and color features~(\cite{opanasenko2024ensemble, cheng2020face}). 
Person re-identification methods, on the other hand, pay attention to the body structure, clothing, color of clothing and accessories, and texture patterns. 
In vehicle re-identification, methods pay attention to the number plate, shape, stickers, and color of the vehicle enabling re-identification from different viewpoints~(\cite{khan2019survey, zhu2023dual}). As an application, traffic surveillance camera systems track vehicles using the above features. In all these ReID scenarios, color features, texture features, and viewpoints are abundant. 
However, the thermal domain is entirely different: images lack color and resolution, texture is less pronounced, and view points are less in number.
Therefore, methods that work well in the RGB domain simply fail with gray-scale thermal images.


Viewpoint is critical in ReID as the feature vectors differ from one viewpoint to another.  In person re-identification, viewpoint invariance can be achieved using large datasets or many viewpoints of each person in the gallery~(\cite{dong2024multi}), a luxury not available in some domains. In vehicle re-identification, such as road traffic and maritime environments, objects (vehicles or vessels) can appear in entirely different views (front, rear, sides, and combinations), unlike in face re-identification---the camera usually points straight at the faces. In the RGB domain, we have the additional advantage of viewpoint-common features such as color and texture patterns, allowing algorithms to lock into these features when re-identifying from multiple viewpoints~(\cite{SPAN, chen2023global}). 
However, in the thermal domain, the absence of common features results in significant differences in the features observed on each side of the vehicles.
Therefore, algorithms should be able to build a feature vector by paying attention to vehicle viewpoints and comparing them with other vehicles using features conditioned by the viewpoints. 
Furthermore, checking which sides are visible in a vehicle image is itself a challenging problem that should be solved prior to the viewpoint-conditioned feature extraction~(\cite{zhou2024dual}).

In this paper, we propose a novel ReID algorithm by constructing viewpoint-conditioned (i.e., agnostic of the viewpoint changes) feature vectors by dividing the foreground image into three segments (front, side, and rear) and extracting features, separately. 
We compare the extracted gallery features with the query feature vector in separate feature spaces and calculate a confidence score based on the viewpoint of the query image. For that, we multiply the L2 distances between the query and gallery features in each space by the corresponding area ratio of those orientations, calculated using the divided foreground image. This approach allows us to compensate for the absence of color and fine features for the shapes of the objects in the thermal domain. We do this by adapting an existing RGB-pre-trained ViT feature extractor in the thermal domain by fine-tuning it to lock into the shapes of the objects (Fig.~\ref{big_picture}). 
While we evaluate our algorithms in both road and maritime environments, the presentation focuses on maritime surveillance to underscore its novelty.
To train and test our algorithms in the thermal maritime domain, we collected our own dataset, which contains videos and images of maritime objects (small and large vessels) from different viewpoints as discussed in Section~\ref{dataset}.

To the best of our knowledge, this is the first time to conduct maritime vessel re-identification in the thermal domain. Our main contributions are as follows:
\begin{itemize}
\item {\bf Viewpoint-conditioned novel ReID algorithm:} 
Constructing a confidence score for gallery images by fusing viewpoint-conditioned features. The algorithm surpasses the SPAN~(\cite{SPAN}) by 16.2\% in mAP in the thermal domain.


\item {\bf A thermal maritime dataset: } Video footage and images of maritime objects from various angles, with COCO annotations~(\cite{coco}), for detection, tracking, and re-identification. To the best of our knowledge, this is the first public dataset in thermal maritime surveillance.\footnote{\textcolor{blue}{https://hevidra.github.io/}} 


\item {\bf Detection and tracking in the thermal domain:} 
Adapting an existing algorithm to the thermal domain by fine-tuning it on the aforementioned dataset, achieving a 61.2\% MOTA score for tracking objects such as vessels, ships, vehicles, and humans in maritime environments and public roads, establishing a baseline.
\end{itemize}

\section{Related Work}\label{sec2}
Re-identification is the process of identifying the same object or individual across different scenes, which typically involves matching features across images or video frames captured at different times and locations. 
It extends into different subdomains including face, human, objects, and vehicle re-identification. 
Face re-identification methods pay attention to specific features such as the iris, dimension of the face, nose, lips, and the color of the eyeball~(\cite{opanasenko2024ensemble, somers2023body, cheng2020face}). 
Due to genetic organizations, each human has a unique combination of these features which makes face ReID possible. 
However, it doesn't facilitate ReID from different angles of the face as the algorithm expects the full frontal view of the face. 
In human ReID tasks, algorithms pay attention to the whole body in addition to the face. Thus, there are other features such as height, shape, body language, and colors of the clothes taken into consideration, allowing recent algorithms to re-identify despite different viewpoints~(\cite{shi2022iranet,bansal2022cloth}). However, in the thermal domain, human ReID has been a challenging task; some approaches get guidance from a visible model to train the thermal model~(\cite{ling2023cross}), while fully thermal approaches suffer from low accuracy~(\cite{yang2023towards}).

In vehicle re-identification, the main challenge is the variability in viewpoint and the high similarity among vehicles of the same model. In literature, there are two main approaches: (i) classify the query image into a single viewpoint and compare the features with gallery images that only belong to that viewpoint. (ii) partition the query image into available viewpoints and compare features extracted from each viewpoint, simultaneously.

For the approach (i), recent work has introduced several techniques that address the viewpoint variation by considering the camera's perspective as a single viewpoint~(\cite{khan2019survey, zhou2018aware, chu2019vehicle}). These methods aim to learn the similarities and differences between images captured from different viewpoints by using loss functions across extracted feature mappings, enabling accurate re-identification across various camera angles. However, these methods have the advantage of color features which are absent in the thermal domain. Furthermore, they have not been tested under poor visibility conditions, such as night-time. In contrast, thermal domain vehicle re-identification based on a single viewpoint is not well explored.~\cite{kamenou2022closing} have proposed a cross-domain model and tried to learn sharable features in both visible and IR domains. The model contains a shareable network followed by two separate streams for two domains increasing the computational complexity. The same authors propose a domain generalization approach for multi-modal vehicle re-identification based on meta-learning~(\cite{kamenou2023meta}) using RGB, near-IR, and IR domains. However, both methods share visible domain features when training, possibly paying less attention to shapes than color features. Furthermore, they expect the query images to be in a close orientation compared to gallery images at the inference, i.e., the method does not force the model to learn orientation-based feature extraction and identity classification. Therefore, we cannot rely on the above methods in the thermal domain due to the lack of discriminative features such as colors and the lack of gallery images from multiple viewpoints. 

For the approach (ii),~\cite{zhou2018vehicle} have proposed a method to learn the features based on the viewpoints by training a CNN to extract eight feature vectors for different viewpoints and compare them with the actual image using the L2 distance as the loss function. 
At the inference, each feature vector is concatenated together and convoluted to create the final multi-view feature vector, which is used for classification. 
SPAN, the ReID algorithm proposed by~\cite{SPAN}, divides the foreground into available viewpoints and extracts and compares features, separately. It uses the area ratio of each viewpoint to weigh the distances in each viewpoint to embed the viewpoint awareness into the algorithm. 
However, since both of the above algorithms are trained in the RGB domain, they have the additional advantage of color features and texture patterns, which are common for multiple viewpoints compared to the thermal domain. 
Therefore, they also pay less attention to the unique shapes of the vehicle when comparing two vehicles of the same model, and rather focus on more discriminative features like colors. 
In contrast, in thermal domain, we do not have color and texture to map features from two orientations, leaving shape as the only option that the model should focus on. Moreover, the above works rely on CNNs for feature extraction, which is less effective at identifying the shapes of objects. Therefore, all of the above methods struggle in the thermal domain for multiple-viewpoint vehicle ReID. 

Object detection and tracking must be performed prior to the re-identification process. Object tracking is the automated process of locating and following objects of interest in images or videos. 
Conventional approaches such as~(\cite{kim2015multiple, tang2017multiple,schulter2017deep}) use two stages for detection and tracking, consuming more computational power and time. In these algorithms, a backbone model is used to detect objects, and then, a separate association algorithm builds tracklets between adjacent frames using those detections. Therefore, they cannot usually be used for real time object tracking due to heavy processing. To overcome these challenges, recent work has moved towards joint detection and tracking approaches, where we detect and track using a single backbone model. Deep SORT~(\cite{wojke2017simple}) compares the appearance of new detections with previously tracked objects within each track to assist data association using a re-identification based approach. 
However, in this method, detection is independently predicted without tracking assistance, which prevents a possible accuracy increment, leading to frequent ID updates of detected objects in occluded or unclear scenarios. 
TraDeS, introduced by Jialian Wu \emph{et al.}~(\cite{wu2021track}), presents an online multi-object tracking algorithm that integrates object detection and tracking to achieve robust and accurate performance using CenterNet~(\cite{duan2019centernet}) as the backbone. Unlike many existing approaches that perform detection independently of tracking, TraDeS incorporates tracking into the detection process to enhance performance in challenging scenarios, including occlusions. It employs a peer-supporting technique where features extracted during detection support the tracking objective, while tracking offsets predicted in the detection stage and features from previous frames enhance current frame detection. This integration not only improves tracking outcomes but also maintains clean tracklets for each detected object, minimizing redundant computations during re-identification. Nevertheless, it has not been tested for the thermal domain, where object tracking is done without color features, specifically with higher frame rates for real-time tracking. 

In this work, we test TraDes for thermal domain for object detection and tracking, and we propose a novel Re-ID technique following the aforementioned approach (ii).

\section{Methodology}\label{sec3}

The framework proposed in this study comprises two subsystems: object tracking and object re-identification, as depicted in Fig.~\ref{big_picture}. The thermal video feed captured by the camera is directed to the object tracking algorithm, which tracks and outputs identified objects. These frames are cropped and forwarded to the re-identification algorithm, which extracts features and compares them with the gallery using a novel viewpoint-conditioned feature comparison technique~(Section~\ref{subsec3.2}). 
The following sections explain each subsystem in detail.

\subsection{Object Tracking for Bounding Box Extraction}\label{Object_track}

\begin{figure}[htbp]
    \centering
    \includegraphics[width=\linewidth]{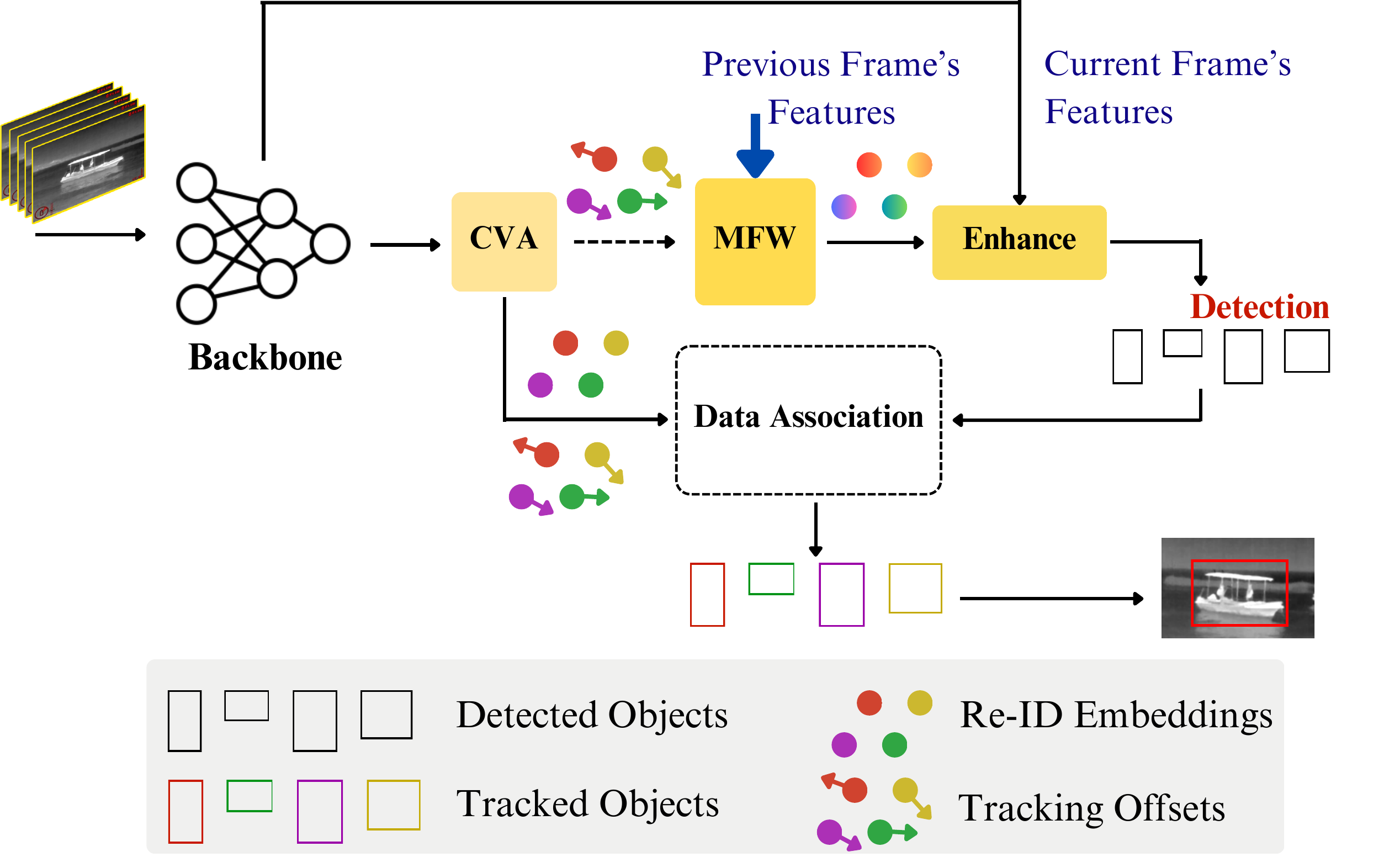}
    \vspace{9pt}
    \caption{Object tracking model architecture: CVA module generates embeddings and derives object motions to improve object tracking accuracy while MFW enhances the feature vector of the current frame based on the tracking history of the past frames. Here, detection and tracking are done using a single backbone, and each part is benefited by the other. Section~\ref{Object_track} discusses the process further.}
    \label{track_picture}
\end{figure}

For object tracking, we adapted TraDeS~(\cite{wu2021track}) algorithm for the thermal domain. First, we use the DLA-34 model~(\cite{yu2018deep}) as the backbone for the feature extraction of input frames. Next, we use two modules to optimize object detection and tracking using the outcomes of each other (Fig.~\ref{track_picture}). The Cost Volume based Association (CVA) module is used to generate embeddings and derive object motions to improve object tracking accuracy. Then, we use the Motion-guided Feature Warper (MFW) module to enhance the object features in the next frame based on the CVA outcomes. More specifically, MFW enhances the feature vector of the current frame based on the tracking history of past frames. This improves the performance of the algorithm, especially when the current frame is occluded. It utilizes tracking cues obtained from the CVA, and propagates them to enhance object features to improve the detection accuracy. Next, object detection is done using CenterNet~(\cite{duan2019centernet}) in the current frame and is associated using a two-round data association technique. In the first round, objects are mapped to the closest tracklet. If it fails, cosine similarity between unmatched tracklet embeddings and the object feature embedding is considered.   This method was chosen because of the model’s ability to integrate detecting, segmenting, and tracking in a single network, which reduces the processing time and improves overall accuracy and efficiency. 
In addition, the model has better real-time tracking of multiple objects compared to previous models discussed in Section~\ref{sec2}, and it is robust to occlusions and appearance changes.


\begin{figure*}[t]
    \centering
    \includegraphics[width=0.8\linewidth]{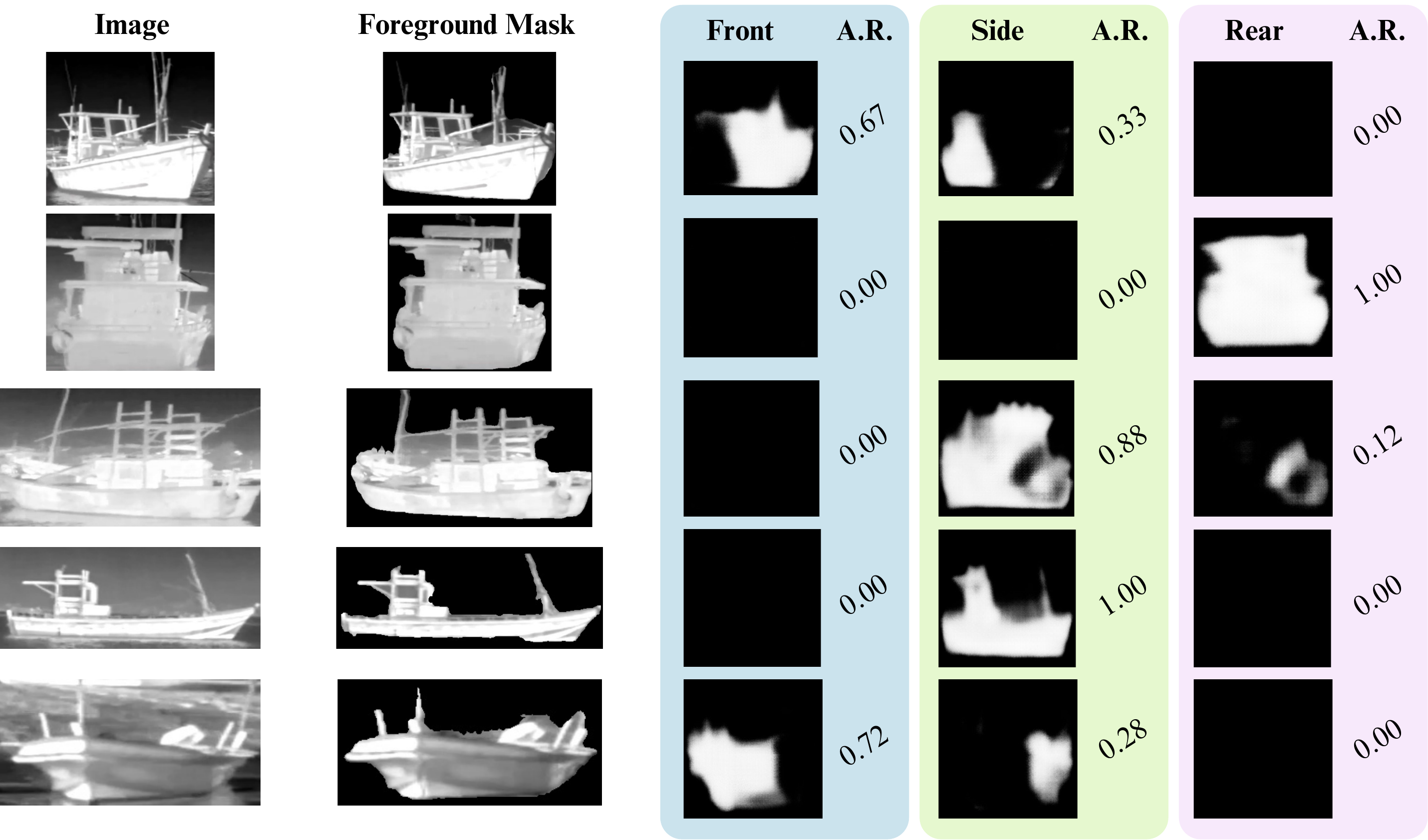}
    \caption{Masks extraction: the second column contains foreground masks generated using the encoder-decoder model, and the following columns contain side masks generated using SPAN. Note that the algorithm is capable of identifying visible sides and assigning accurate values to area ratios. The process is shown in Fig.~\ref{fig_reid_arch}.(b), as a part of the complete algorithm.}
    \label{fig_enc_dec}
\end{figure*}

\subsection{Vehicle Re-identification} \label{subsec3.2}

In vehicle ReID, our target is to extract the identity of a given query image using a set of gallery images. 
Here, the main challenges are the lack of available vehicle data in thermal domain and the change of features for each vehicle with different camera viewpoint. 
To tackle these issues, we used a model robust to the viewpoint, which can generalize well with a small amount of data. 
As the first step, we mask out the foreground (the vehicle) from a given frame. Since thermal images do not contain color features and the intensity distribution of the foreground and the background are similar, conventional algorithms such as GrabCut~(\cite{rother2004grabcut}) do not perform well in this task. As a solution, we used an encoder-decoder architecture (Appendix: Fig~\ref{enc_dec_archt}) with residual connections to build the foreground mask of a given frame. 
We annotated foreground masks of 300 images as ground truth labels, and trained the model using those frames as input. 
Then, as shown in Fig.~\ref{fig_enc_dec}, the trained model was tested on previously unseen data, demonstrating its capability to accurately mask out foreground elements with complex viewpoints and intensity variations. Therefore, we propose this encoder-decoder architecture as a foreground extractor, specifically for colorless images, given that it can be trained on such data.

\begin{figure*}[t]
    \centering
    \includegraphics[width=0.85\linewidth]{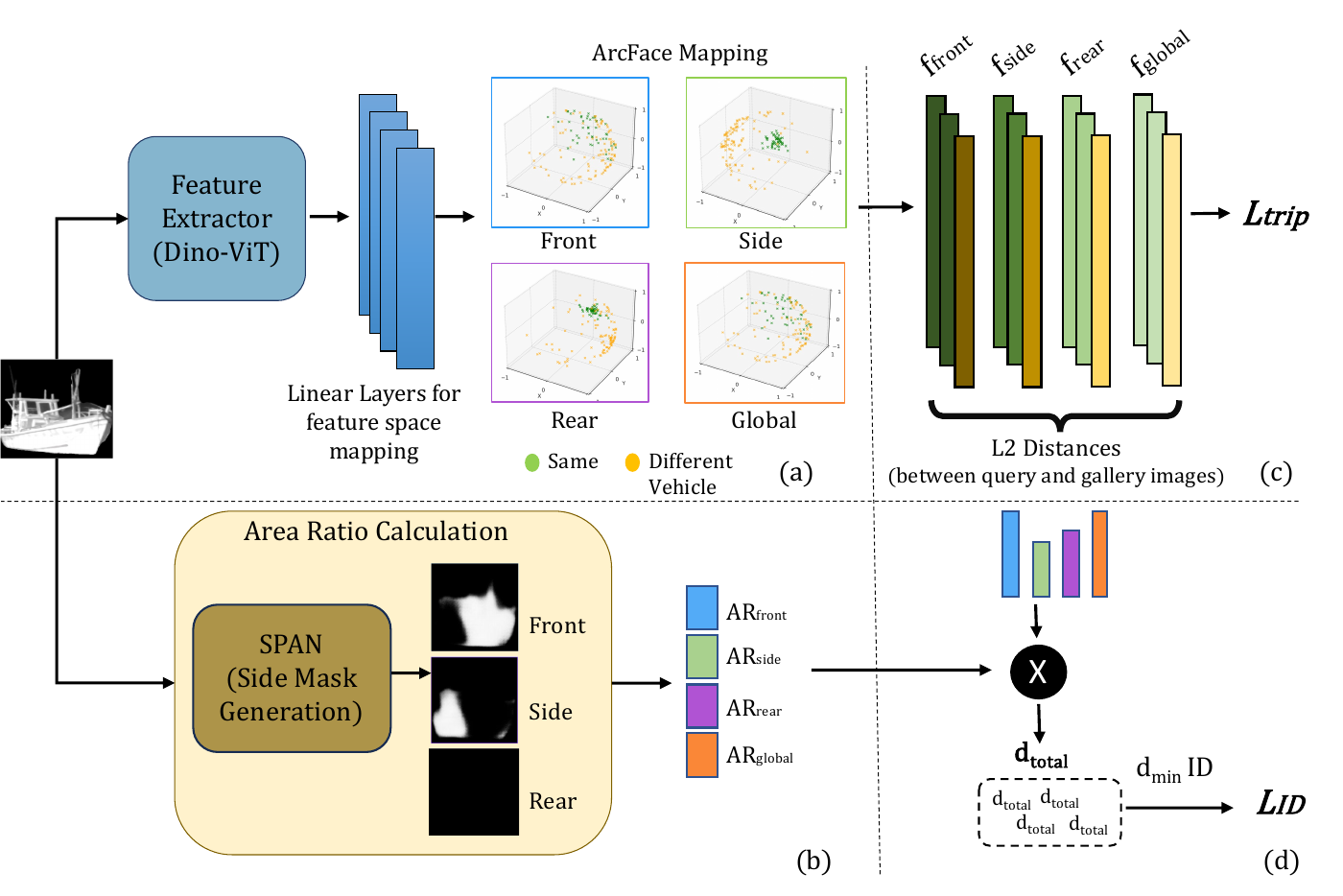}
    \caption{Re-identification subsystem. (a) extracts features using the vision transformer and maps them to four latent spaces. (b) calculates the area ratios of each visual side of the vehicle. (c) and (d) calculate triplet loss and the ID loss, respectively.}
    \label{fig_reid_arch}
\end{figure*}

\begin{figure*}[t]
    \centering
    \includegraphics[width=0.85\linewidth]{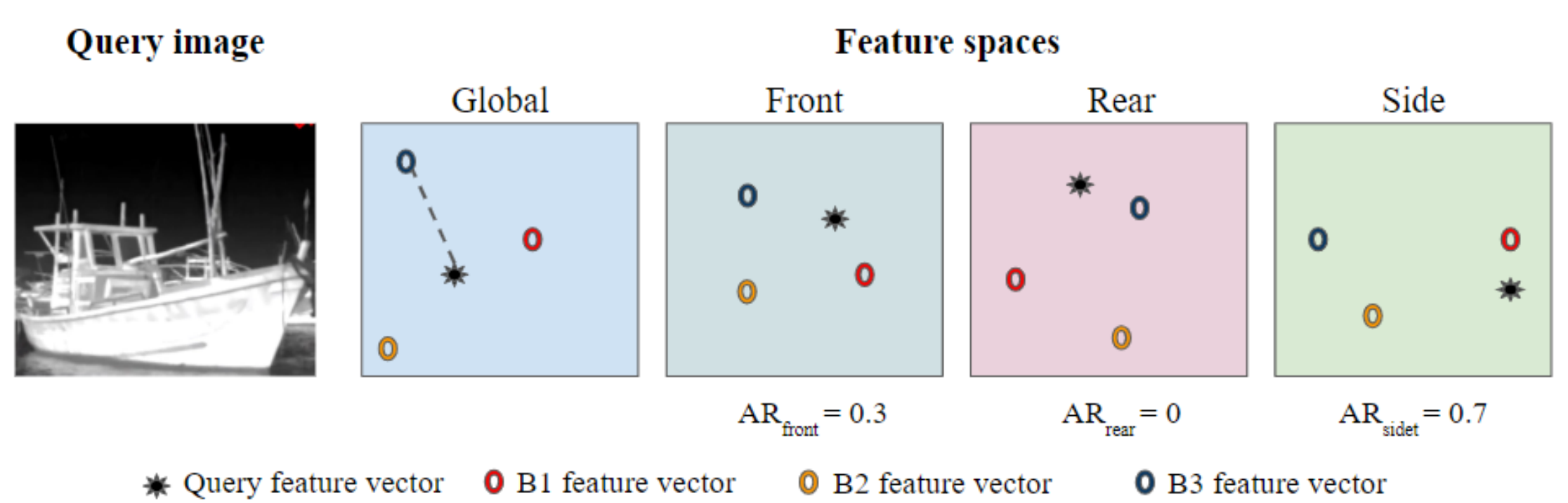}
    \caption{L2 distance calculation in separate spaces. According to the given vehicle, more attention should be paid to distances in the side space while completely neglecting the rear space. Since B1 is the closest in the side space and comparatively close in the front space, the best match will be B1 for the given query image.}
    \label{fig_feature_spaces}
\end{figure*}

Next, the extracted foreground is fed to the ReID model, where we use an architecture as shown in Fig.~\ref{fig_reid_arch}.(a) to extract the features of a given vehicle. As the feature extractor, we used the pre-trained Dino ViT transformer model presented in \cite{caron2021emerging}. As shown in the Fig.~\ref{fig_reid_arch}.(a), four parallel linear layers are used to map the extracted feature vector to four latent spaces, namely global, front, rear, and side. 
We use these latent spaces to train the model to recognize vehicle identities in different viewpoints since the vehicle's features drastically change with the viewpoint. 
To get a better feature distribution, ArcFace~(\cite{deng2019arcface}) mapping is used in each space. ArcFace is a feature analyzing technique that maps feature vectors onto a hypersphere, enhancing discrimination between different identities by maximizing inter-class variance while minimizing intra-class variance. Next, we calculate L2 distances to each vehicle in the gallery, in each space, as shown in Fig.~\ref{fig_reid_arch}.(c). Inspired by the SPAN model, these distances are multiplied by area ratios~(Fig.\ref{fig_reid_arch}.(b)) of each viewpoint, as calculated in~eq.~(\ref{eq:area_ratio}), to embed the viewpoint information to the result. To calculate the area ratios, we use a decoder architecture to break down the foreground to side-wise masks (front, side, and rear) and calculate the ratio of white pixels in the side mask over the foreground (Fig.~\ref{fig_enc_dec}). We calculate the total distance, the weighted sum of feature distances in each space, to a gallery image as given in eq.~(\ref{eq:1}).
This equation handles self-occlusion scenarios and also eliminates the influence of distance comparisons in non-visible viewpoints of a query image by applying an area ratio, which becomes zero for such cases (Fig.~\ref{fig_feature_spaces}).


\begin{equation}
\label{eq:area_ratio}
\begin{aligned}
\mathrm{AR_{side X}} &= \frac {\text{Area of the sideX view mask}}{\text{Area of the foreground mask}}, \\
& \text{where sideX} \in \{\text{front}, \text{side}, \text{rear}\}.
\end{aligned}
\end{equation}

\begin{equation}
\label{eq:1}
\begin{aligned}
& d_\mathrm{total}(\mathrm{ID,Image}) = \quad \left[ d_\mathrm{global}(\mathrm{ID,Image}) \right. \\
& \quad + d_\mathrm{front}(\mathrm{ID,Image}) \cdot \mathrm{AR_{front}} \\
& \quad + d_\mathrm{side}(\mathrm{ID,Image})\cdot \mathrm{AR_{side}} \\
& \quad + d_\mathrm{rear}(\mathrm{ID,Image}) \cdot \left.\mathrm{AR_{rear}} \right] /2
\end{aligned}
\end{equation}

When training the re-identification model, we freeze the area ratio calculation and fine-tune the linear layers to map the extracted features to the four spaces. We use identity classification (Section~\ref{lab:Identity Classification Loss}) and triplet loss (Section~\ref{lab:Triplet Loss}) in the training process. 



\subsubsection{Identity Classification Loss ($L_{ID}$)}
\label{lab:Identity Classification Loss}
In the identity classification, we calculate the class label using eq.~(\ref{eq:1}), where the identity is given as the ID of the gallery image with the minimum distance. Then, we use the cross-entropy loss as the identity classification loss.

\subsubsection{Triplet Loss ($L_{trip}$)}
\label{lab:Triplet Loss}
 To promote discrimination and effective feature learning, we use triplet loss with Euclidean distance on the four feature spaces, separately. For the negative and positive samples, we manually use sample thermal images for each vehicle identity to make sure the model is robust to different viewpoints. The total loss is calculated as in eq.~(\ref{eq:2}),
\begin{equation}
\label{eq:2}
\begin{aligned}
L_{\mathrm{Total}} = \lambda_{\mathrm{ID}} \cdot L_{\mathrm{ID}} +  \lambda_{\mathrm{Triplet}} \cdot L_{\mathrm{Triplet}}
\end{aligned}
\end{equation}
where  $\lambda_{\mathrm{ID}}$ and  $\lambda_{\mathrm{Triplet}}$ are hyper parameters empirically selected.

\section{Experiments}\label{sec4}

In this section, we describe datasets used, experimental methods and the procedure followed. We used a GeForce RTX 2080 GPU to infer the system and report the performance indicators mentioned in Table~\ref{Table:Indicators}. We conducted extensive experiments on multiple datasets using several state-of-the-art methods along with our method and our dataset.

\begin{table}[h]
\caption{Performance Indicators used to evaluate each algorithm}
\centering
\footnotesize
\begin{tabular}{@{}p{4cm}r@{}}
\toprule
\textbf{Algorithm} & \textbf{Indicator} \\
\midrule
Object detection and tracking & MOT, IDF1, fps \\
Re-identification & Top 1, Top 5, mAP \\
\bottomrule
\end{tabular}
\label{Table:Indicators}
\end{table}

\begin{table}[h!]
\caption{Experiment catalog}
\centering
\footnotesize
\begin{tabular}{@{}p{2cm}p{1.3cm}r@{}}
\toprule
\textbf{Subsystem} & \textbf{Methods} & \textbf{Datasets used}\\
\midrule
Object tracking & TraDeS & Ours, MOT17, SMD  \\
Re-identification & Our, SPAN & Ours, VeRi776, VehicleID, VesselID-539 \\
\bottomrule
\end{tabular}
\label{Table:Datasets}
\end{table}

\begin{table*}[t!]
    \centering
    \footnotesize
    \caption{Results of the TraDes algorithm on different domains, evaluated using MOT17, SMD, and Our dataset. Note that the algorithm has obtained a 61.2\% MOTA score in the IR domain, which is almost the same as the RGB domain performance. Therefore, the domain adaptation is successfully achieved while conserving the performance of the algorithm.}
    \vspace{1mm}
        \begin{tabularx}
        {0.7\textwidth}{@{}l l l X X X X X X r@{}}
            \toprule
            Dataset & Domain & MOTA$\uparrow$ & IDF1$\uparrow$ & MT$\uparrow$ & ML$\downarrow$ & FP$\downarrow$ & FN$\downarrow$ & IDS$\downarrow$ & FPS$\uparrow$ \\
            \midrule
            MOT17 & RGB & 63.5 & 67.7 & 36.3 & 21.5 & 4.5 & 31.4 & 3147 & 30 \\
            MOT17 & B\&W & 58.8 & 64.6 & 31.6 & 29.5 & 3.3 & 37.3 & 3524 & 30 \\
            SMD & Near-IR & 59.5 & 65.2 & 33.1 & 28.1 & 3.4 & 36.6 & 2831 & 30 \\
            \textbf{Our dataset} & \textbf{IR} & \textbf{61.2} & \textbf{65.4} & \textbf{35.5} & \textbf{24.9} & \textbf{3.6} & \textbf{33.2} & \textbf{2109} & \textbf{30}\\
            \bottomrule
        \end{tabularx}
    \label{Table:TraDeS}
\end{table*}

\subsection{Datasets}\label{dataset}

\textbf{Our dataset}: Our maritime dataset, captured using a FLIR M232 marine thermal camera, contains videos of moving vessels and maritime objects which are suitable for testing detection and tracking algorithms in the thermal domain. We drew bounding boxes for 4 classes: vessels, ships, humans, and jet skies. Furthermore, the dataset contains images of 40 small vessels and 32 large vessels from different viewpoints that can be used to train re-identification algorithms. It contains annotated video feeds of swimming and possible human trafficking activities that can be modeled as suspicious activities.\

\textbf{VeRi776~(\cite{Veri776})}: This RGB dataset is a comprehensive collection of vehicle re-identification data, comprising a total of 49,357 images that feature 776 distinct vehicles captured by 20 different cameras. Furthermore, it contains bounding boxes and information regarding vehicle types, colors, and brands. 

\textbf{VesselID-539~(\cite{qiao2020marine})}: This RGB dataset is a collection of marine vehicle images that were sourced from the website Marine Traffic (www.marinetraffic.com). The raw vehicle image dataset encompasses a substantial quantity of data, comprising over 149,465 images representing 539 distinct vessels. On average, each vehicle in the dataset is represented by approximately 277 images. 

\textbf{VehicleID~(\cite{b14})}: VehicleID dataset contains 26,267 RGB images of vehicles captured from different viewpoints in the daytime. For our experiments, we used 500 identities as the training and validation dataset and another 250 identities as the query and gallery images.

\textbf{Singapore Maritime Dataset (SMD)~(\cite{b15})}: The Singapore Maritime Dataset consists of meticulously curated high-definition near-IR videos captured using strategically positioned Canon 70D cameras around the waters of Singapore. It encompasses on-shore videos acquired from fixed platforms along the shoreline, as well as on-board videos captured from moving vessels, providing diverse perspectives of the maritime environment. This division ensures comprehensive coverage and enables analysis across various viewpoints and scenarios. 

{\textbf{RGBNT100~(\cite{rgbn100})}: The dataset contains 17250 IR images of 100 vehicles captured in 8 different viewpoints. Furthermore, the dataset contains the RGB and near-IR images of the same vehicles. The resolution is $640\times 480$ for thermal images and $1920\times 1080$ for RGB images. For our experiments, we consider only the thermal images.


Table~\ref{Table:Datasets} summarizes the datasets and methods used for evaluation purposes in the results in Section~\ref{sec5}.

\subsection{Evaluation Matrics}

To evaluate the object detection and tracking algorithm, we use the common 2D MOT evaluation metrics~(\cite{bernardin2008evaluating}): Multiple-Object Tracking Accuracy (MOTA), ID F1 Score (IDF1), the number of False Negatives (FN), False Positives (FP), times a trajectory is Fragmented (Frag), Identity Switches (IDS), and the percentage of Mostly Tracked Trajectories (MT) and Mostly Lost Trajectories (ML).

For re-identification tasks, we use the Top1, Top5, and Mean Average Precision (mAP). Top1 represents the accuracy of the model in predicting the correct class label as the highest-ranked prediction and Top5 score is the percentage of times the true label is included in the top 5 predictions. mAP evaluates the precision-recall tradeoff by measuring the model's ability to rank predictions accurately and capture relevant instances.

\subsection{Training} \label{sec4.2}

{\bf For object detection and tracking},  we trained TraDeS for 5 classes, including vessels, cars, ships, humans, and jet skies. The training was done in two phases. In the first phase, we trained on a subset of classes in the COCO dataset (RGB), relevant to the specific use case, such as vehicles and humans. We converted RGB data to grayscale to make COCO images more similar to thermal images. In the second phase, we completely moved to the thermal domain by tuning the model using SMD dataset~(\cite{b15}), along with our thermal data. Subsequently, we fine-tuned the algorithm by adjusting the hyperparameters,  learning rate, and detection threshold.

{\bf In the re-identification module} as shown in Fig.~\ref{big_picture}(b), we trained area ratio calculation and the feature mapping parts, separately. For area ratio calculation, we train the model responsible for part attention in SPAN using the reconstruction loss used in the original work~\cite{SPAN} and our thermal dataset. In the feature mapping part, we used Dino-ViT, which is pre-trained on the ImageNet dataset, for the initial feature extraction. Then, we used transfer learning to train the linear layers in Fig.~\ref{fig_reid_arch}.(a) using our thermal image dataset while keeping the area ratio calculation in the inference as it is already trained. We use triplet loss to learn the discriminative features of vessels and ID loss to learn the distance matrix for a correct classification of a given query image.

\section{Results and Discussion}\label{sec5}

In this section, we first evaluate the performance of the TraDes algorithm in the thermal domain for vehicle, human, and maritime object tracking to show that it can obtain similar results as in the RGB domain after the adaptation that we introduced. 
Second, we show that the viewpoint-conditioned re-identification approach proposed in this paper outperforms SPAN method in both RGB and thermal domains obtaining higher mAP values. 
Then, as ablations, we present results with and without viewpoint-conditioned feature comparison, and effects of CNN and ViT based feature extractions. Finally, we evaluate the effect of the number of viewpoints in the feature comparison for the final outcome. 


\begin{figure*}[t]
    \centering
    \includegraphics[width=\linewidth]{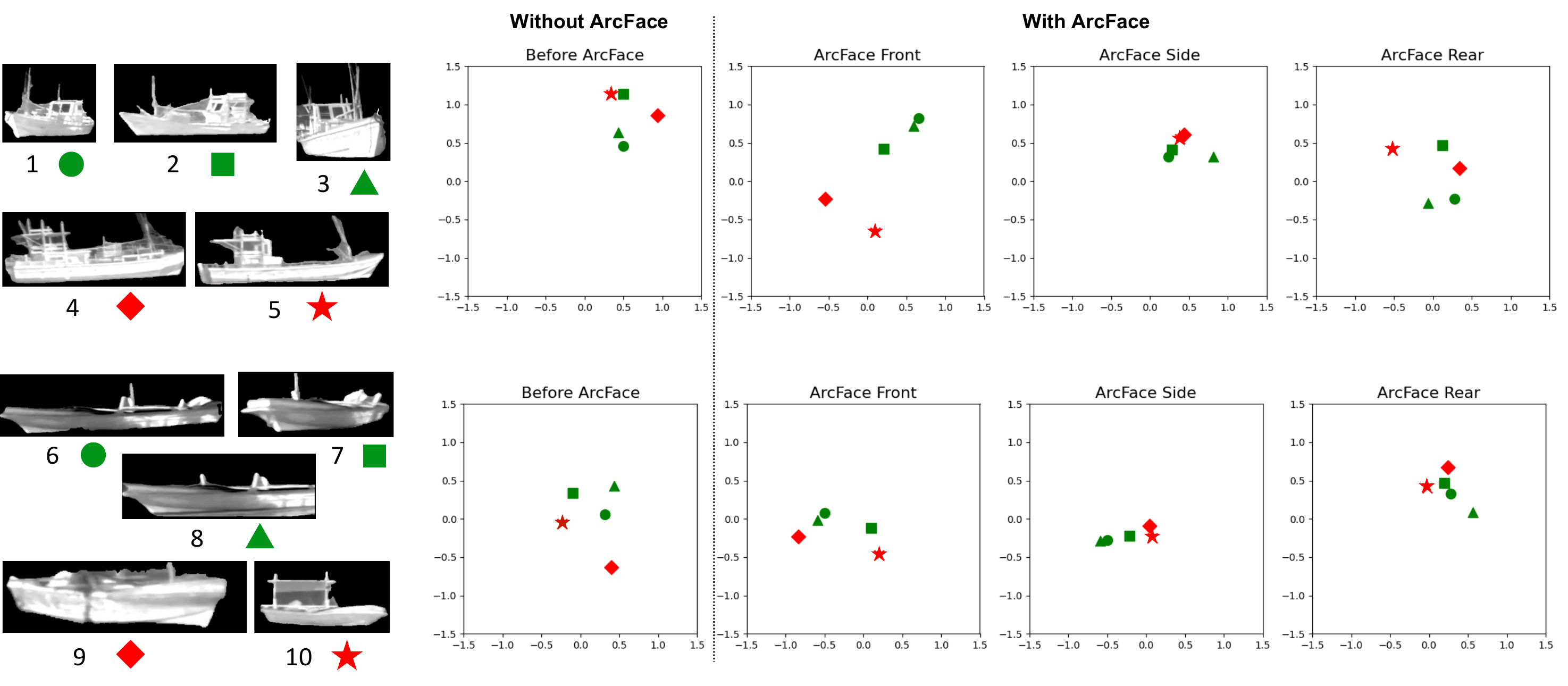}
    \caption{Visual representation of the feature mappings without ArcFace and with ArcFace using dimensionality reduction of feature vectors. Here, green color markers (1, 2, 3, 6, 7, 8) represent images of the same vessel from different angles, while red markers (4, 5, 9, 10) represent images of other vessels. Note that feature representations of similar vessels get closer in corresponding spaces according to their orientations, while different vessels are separated by a considerable margin.}
    \label{reid_visual}
\end{figure*}

\begin{table*}[h!]
\footnotesize
\centering
\caption{Performance comparison of ReID algorithms. Our algorithm outperforms SPAN in both RGB and thermal domains, obtaining the highest mAP score across all datasets. In particular, in the thermal domain that we focus, our results well-surpass existing results. VC: viewpoint-conditioned.}
\vspace{2mm}
\begin{tabular}{p{1.6cm}p{0.9cm}p{1.2cm}|p{0.8cm}p{0.65cm}p{0.65cm}|p{0.65cm}p{0.65cm}p{0.65cm} | p{0.65cm}p{0.65cm}r@{}}
\toprule
\multicolumn{3}{c}{} &
\multicolumn{3}{>{\centering\arraybackslash}c}{\textbf{SPAN}} &
\multicolumn{3}{>{\centering\arraybackslash}c}{\textbf{ViT Base}} &
\multicolumn{3}{>{\centering\arraybackslash}c}{\textbf{ViT Base + VC (Ours)}} \\
{\bf Dataset} & Domain & Num Cls &{\bf Top1} & {\bf Top5} & {\bf mAP} & {\bf Top1} & {\bf Top5} & {\bf mAP} & {\bf Top1} & {\bf Top5} & {\bf mAP}\\
\midrule
VehicleID & RGB & 250 & 78.22 & 85.16 & 56.10 & 71.40 & 82.00  & 52.02 & \textbf{78.40} & \textbf{86.80}  & \textbf{57.62}\\ 

VeRi776 & RGB & 200 & \textbf{94.0} & \textbf{97.6} & \textbf{68.9}  & 80.16 & 85.16  & 58.91  & 91.67 & 94.67  & 67.26 \\ 

VesselID-539 & RGB & 200 & 82.43 & 86.67 & 58.10 & 82.52 & 86.70  & 60.18 & \textbf{82.60} & \textbf{86.93}  & \textbf{61.44}\\

\midrule
RGBN100 & IR & 50 & 75.13 & 81.81 & 53.65  & 82.11 & 86.24  & 60.81 & \textbf{84.44} & \textbf{88.51}  & \textbf{64.26}\\

\textbf{Our datatset} & IR & 35 & 78.37 & 84.43 & 55.88  & 79.21 & 84.50  & 61.33 & \textbf{81.82} & \textbf{86.36}  & \textbf{63.08}\\
\bottomrule
\end{tabular}
\vspace{-3mm}
\label{Table:reid_results}
\end{table*}

\begin{table*}[h!]
\footnotesize
\centering
\caption{Performance comparison with the number of viewpoints considered in the feature comparison using our algorithm (ViT Base + VC). Including more than one view consistently improves the performance both in RGB and thermal domains. The highest improvement is in the thermal domain.}
\vspace{2mm}
\begin{tabular}{@{}p{2cm}p{1.2cm}p{1.2cm}|p{0.9cm}p{0.9cm}p{0.9cm}|p{0.9cm}p{0.9cm}r@{}}
\toprule
\multicolumn{3}{c}{} &
\multicolumn{3}{>{\centering\arraybackslash}c}{\textbf{Largest view}} &
\multicolumn{3}{>{\centering\arraybackslash}c}{\textbf{All views}} \\
{\bf Dataset} & Domain & Num Cls & {\bf Top1} & {\bf Top5} & {\bf mAP} & {\bf Top1} & {\bf Top5} & {\bf mAP}\\
\midrule
VehicleID & RGB & 250 & 71.80 & 82.40  & 52.76 & \textbf{78.40} & \textbf{86.80}  & \textbf{57.62}\\ 

VeRi776 & RGB & 200 & 78.83 & 84.83  & 54.56  & \textbf{91.67} & \textbf{94.67}  & \textbf{67.26} \\ 

VesselID-539 & RGB & 200 & 72.33 & 78.56  & 52.47 & \textbf{82.60} & \textbf{86.93}  & \textbf{61.44}\\  
\midrule

RGBNT100 & IR & 50 & 78.21 & 81.02  & 59.11 & \textbf{84.44} & \textbf{88.51}  & \textbf{64.26}\\

\textbf{Our dataset} & IR & 35 & 68.18 & 72.72  & 50.75 & \textbf{81.82} & \textbf{86.36}  & \textbf{63.08}\\
\bottomrule
\end{tabular}
\vspace{-3mm}
\label{Table:view_points}
\end{table*}

{\bf Evaluation of our tracking algorithm}:  We evaluated the performance of the TraDes algorithm in the near-IR and IR domains using SMD and our dataset. As shown in Table~\ref{Table:TraDeS}, higher MOTA and MAP scores in our dataset clearly indicate that the algorithm has successfully adapted to the specified classes (vessels, ships, vehicles and humans) in the IR domain. The algorithm has obtained a 61.2\% MOTA score in the IR domain, which is almost the same as the RGB domain performance. It indicates that we can track objects without color features with only a minor drop in the performance indicators. Also, we could maintain a 15 fps processing speed which is suitable for real-time online tracking. 
Therefore, the domain adaptation has been successfully achieved while conserving the performance of the algorithm.

{\bf Evaluation of our re-identification algorithm}: 
Our re-identification algorithm convincingly surpasses SPAN and ViT baselines in the IR domain while showing competitive performance even in the RGB domain with a higher mAP score.
 In our dataset, it achieved a Top1 accuracy of 81.82\% and a mAP score of 63.08\% compared to SPAN's Top1 accuracy of 78.37\% and mAP score of 55.88\%, indicating the effectiveness of our method in handling thermal images with multiple vehicle viewpoints (4.5\% increment in the Top1 score in Table~\ref{Table:reid_results}). 
 Furthermore, the algorithm achieved 12.4\% increment in the Top1 score and 19.8\% increment in the mAP score for the RGBNT100 dataset (contains thermal vehicle images) generalizing the higher performance for vehicle re-identification. On average, our algorithm outperforms SPAN by 8.4\% and 16.3\% in Top1 and mAP score in the thermal domain. To extend the generalizability of our method, we conducted experiments with above mentioned RGB datasets and showed that our method works competitively in the RGB domain, as well. In the VesselID-539  dataset, our algorithm achieved a Top1 accuracy of 82.6\% (compared to SPAN's 82.43\%), indicating that the proposed method, which is specified for thermal domain performance, is robust in the RGB domain, as well. Moreover, across all datasets, our method shows considerably higher mAP scores, outperforming SPAN by 7.7\% on average.

We explain the gain in the results using two concepts: (1) The ViT can extract complex features using its attention mechanism which enables paying more attention to specific shapes and masks of vehicles/vessels (Fig.~\ref{reid_visual}). Using the triplet loss and the ID loss mentioned in Section~\ref{subsec3.2}, it learns to put more weight on critical features of the shape of the vehicle/vessel, eventually pushing feature vectors of similar objects closer in the feature space. As a result, the mAP score increases as shown in the third main column (ViT Base) of Table~\ref{Table:reid_results}. However, the ViT only cannot maintain a good Top1 score due to the vast variations of the orientation. (2) The ArcFace mapping further organizes features in 3 separate spaces (front, side, and rear). It increases the inter-class distances in each side-space and enables side-wise feature comparison, increasing the Top1 and Top5 scores of the algorithm, as shown in the third (ViT Base) and fourth (ViT Base + VC) columns of  Table~\ref{Table:reid_results}, which serves as an ablation study. Furthermore, our algorithm is capable of maintaining a higher mAP score even with a higher number of classes and different orientations compared to the query image, due to the space-separated feature comparison.  SPAN, in contrast, loses accuracy with increasing orientation changes, resulting in low mAP scores. As shown in Fig.~\ref{reid_visual}, vessels 1, 2 and 3 belong to the same vehicle ID, and 4 and 5 belong to different vessels. However, without ArcFace mapping (column 1), vessels 2 and 5 lie closer, suggesting they belong to the same class. When we consider the side space of the ArcFace mapping, they have been placed apart, suggesting that vessels 1 and 2 belong to the same vehicle ID, which is correct. Furthermore, vessels 6 and 8, which belong to the same vehicle ID and have a more side orientation, are closer in the side space, while vessels 7 and 9, which belong to different IDs, are pushed apart in the front and side spaces. Therefore, using ArcFace mapping, we can achieve higher accuracy by comparing feature vectors in relevant ArcFace spaces according to the vehicle orientation, which is done by the area ratio multiplication at the end (Fig.~\ref{fig_reid_arch}.(b)).

Moreover, we did another ablation study on the number of viewpoints considered in the feature comparison. Here, we considered the side with the highest area ratio as the largest view and did the feature comparison only for that side. As shown in Table~\ref{Table:view_points}, feature comparison in multiple viewpoints (typically 2 views appear in an image) increases the Top1 score by 15\%, on average across all the datasets. This increment comes from the features extracted from the minor sides visualized in the vehicle/vessel image, which can be critical in some situations.


\section{Conclusion}\label{sec6}

In this paper, we have proposed a novel re-identification algorithm for vehicle and vessel re-identification for thermal domain using a viewpoint-conditioned feature selection technique. 
It computes the final confidence score for a query image by dividing features into front, side, and rear spaces, comparing them with gallery feature vectors, and conditioning on viewpoint area ratios. We surpass the state-of-the-art methods by 19.7\% and 12.8\% in mAP scores for the RGBNT100 and a thermal maritime dataset acquired by us, respectively. To the best of our knowledge, this will be the first public dataset in thermal domain which contains video footage and images of maritime objects from various angles, with COCO annotations, for detection, tracking, and re-identification. Furthermore, we adapted a tracking algorithm to the thermal domain by fine-tuning it on the aforementioned dataset, achieving a 61.2\% MOTA score for tracking objects such as vessels, ships, vehicles, and humans, establishing a baseline. As future work, we plan to explore the re-identification of occluded vehicles/ vessels with a higher number of gallery IDs.



\section{Acknowledgment}\label{ack}

Funding for the FLIR thermal camera was provided by the Senate Research Committee Capital Grant: SRC/CAP/2018/02. Computational resources were provided by the Creative Software Pvt. Ltd.

\bibliographystyle{plainnat}
\bibliography{manuscript}

\clearpage
\newpage

\appendix
\section*{Appendix}

To compare the performance of GrabCut and the encoder-decoder model, we extract the foreground of 5 images as shown in Fig~\ref{grabcut}. Since the original image (output of the tracking algorithm) contains a tight bounding box around the vessel, we chose the initial rectangle as the complete image in the GrabCut method. The encoder-decoder model is trained on some other vessel data, and therefore, these images are unseen. According to the results, our encoder-decoder method has better performance compared to GrabCut especially when dealing with small extensions in the vessel. Furthermore, GrabCut struggles to remove the sea parts when a glare is present due to reflections.

 Fig~\ref{enc_dec_archt} contains the architecture of the encoder-decoder model we proposed for the foreground extraction for thermal images.

\begin{figure}[h!]
    \centering
    \includegraphics[width=\linewidth]{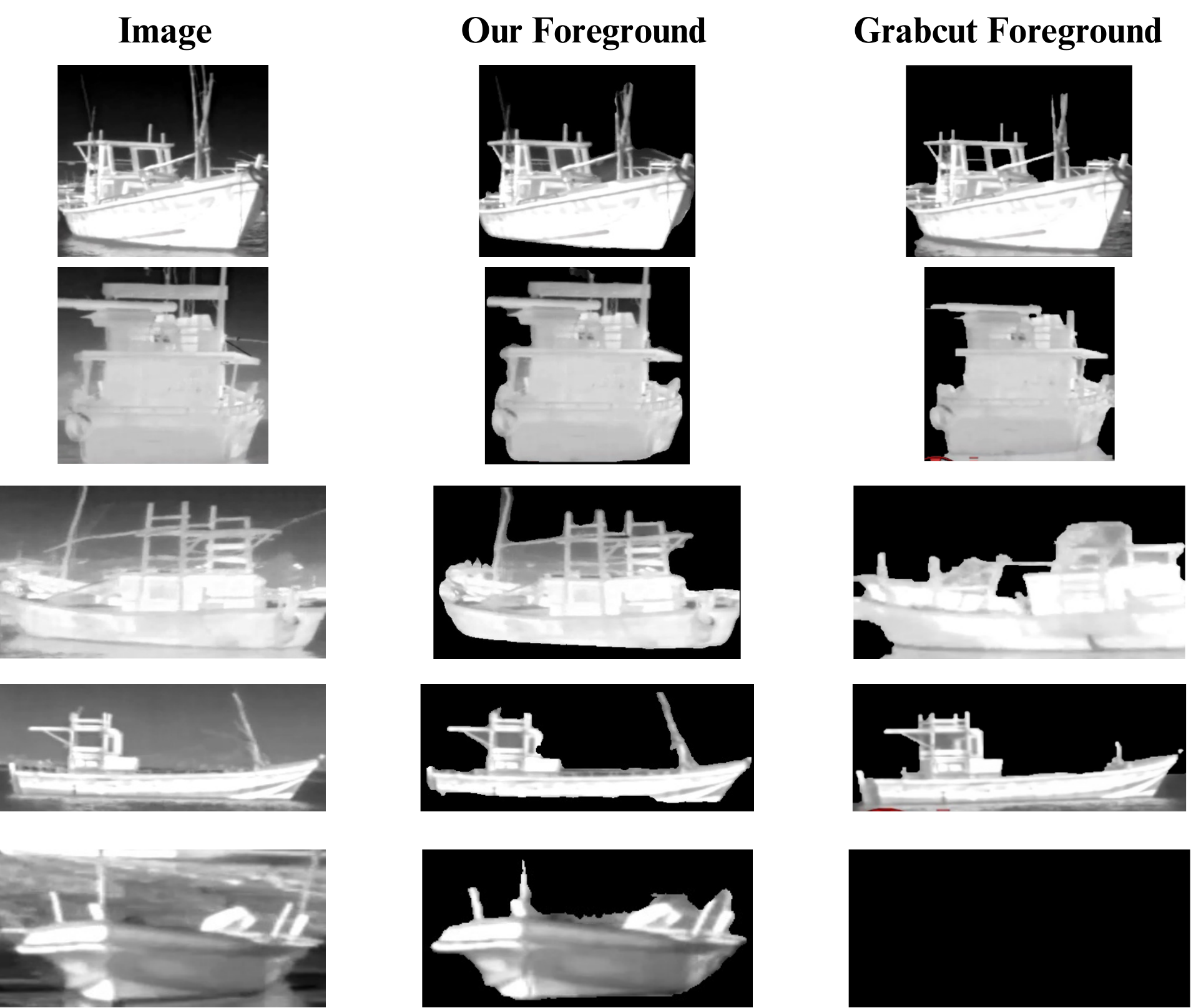}
    \caption{Model architecture of the encoder-decoder model proposed for foreground extraction in thermal images.}
    \label{grabcut}
\end{figure}

\begin{figure}[h!]
    \centering
    \includegraphics[width=0.82\linewidth]{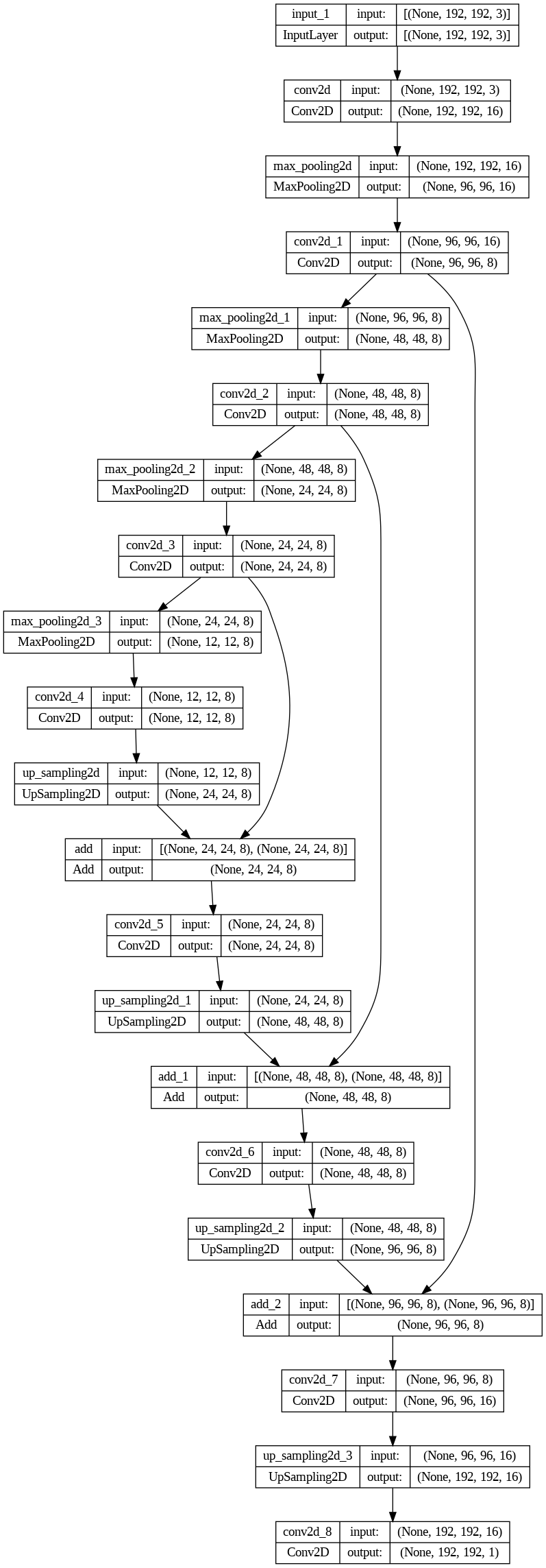}
    \caption{Visual performance comparison between GrabCut and our method in foreground extraction in thermal domain.}
    \label{enc_dec_archt}
\end{figure}

\clearpage

\end{document}